\documentclass[conference]{IEEEtran}
\IEEEoverridecommandlockouts
\usepackage{cite}
\usepackage{amsmath,amssymb,amsfonts}
\usepackage{algorithmic}
\usepackage{graphicx}
\usepackage{textcomp}
\usepackage{xcolor}
\def\BibTeX{{\rm B\kern-.05em{\sc i\kern-.025em b}\kern-.08em
    T\kern-.1667em\lower.7ex\hbox{E}\kern-.125emX}}
\begin{document}

\title{Self Optimisation and Automatic Code Generation by Evolutionary Algorithms in PLC based Controlling Processes\\
{\footnotesize Evaluation approach for applications and processes in the industrial sector}
}

\author{\IEEEauthorblockN{1\textsuperscript{st} Marlon Löppenberg}
\IEEEauthorblockA{\textit{Automation Technology and Learning Systems} \\
\textit{South Westphalia, University of Applied Sciences}\\
Soest, Germany \\
loeppenberg.marlon@fh-swf.de}
\and
\IEEEauthorblockN{2\textsuperscript{nd} Andreas Schwung}
\IEEEauthorblockA{\textit{Automation Technology and Learning Systems} \\
\textit{South Westphalia, University of Applied Sciences}\\
Soest, Germany \\
schwung.andreas@fh-swf.de}

}

\maketitle

\begin{abstract}
The digital transformation of automation places new demands on data acquisition and processing in industrial processes.
Logical relationships between acquired data and cyclic process sequences must be correctly interpreted and evaluated.
To solve this problem, a novel approach based on evolutionary algorithms is proposed to self optimise the system logic of complex processes.
Based on the genetic results, a programme code for the system implementation is derived by decoding the solution.
This is achieved by a flexible system structure with an upstream, intermediate and downstream unit.
In the intermediate unit, a directed learning process interacts with a system replica and an evaluation function in a closed loop.
The code generation strategy is represented by redundancy and priority, sequencing and performance derivation. 
The presented approach is evaluated on an industrial liquid station process subject to a multi-objective optimisation problem.
\end{abstract}


\section{Introduction}
Production plants are used in many areas of the manufacturing and automation industry.
In recent years, the process automation sector has undergone a transformation towards digitalisation and intelligent networking, referred to as Industry~$4.0$~\cite{Kagermann}.
With the reinterpretation of automation processes, more and more digital processing is moving to the field level, such as image processing for cameras~\cite{Khan} or interface extensions for wireless communication for the Internet of Things (IoT) applications~\cite{Chaudhary}\cite{Zarzo}. 
At this level, all the data from sensors and actuators is processed and evaluated centrally by a Programmable Logic Controller (PLC).
The code for the PLC is currently written by a programmer who needs to understand and know the process in detail. 
Depending on the size of the solution to be implemented, this task is very extensive and time-consuming.
In addition, the programme structure developed is subject to a high degree of variability and depends on the skills of the programmer. 
An improvement would be to solve these problems with an automated process.\\

In view of the problem described, a self optimising evaluation process with automatic code generation is proposed. 
A suitable method for this implementation is an Evolutionary Algorithm (EA).
It combines situational understanding with the optimisation of different performance objectives on a stochastic and metaheuristic basis.
To generate a solution approach, the optimisation setup interacts with a Digital Twin (DT)~\cite{Singh} in a cyclical evaluation.
The problem is considered as a Multi-Objective Optimisation Problem (MOOP)~\cite{Kalyanmoy} resulting in a Pareto front in the objective space. 
By separating the optimisation from the dynamic system, decoupling is achieved and the process is split by the system setup into an active and passive part.
The flexible interface definition of the system setup determines the interaction with the real system.
A fitness function is implemented to control the optimisation based on the system specific problem. 
This function can also be defined with specific parameters in order to influence the optimisation.
The presented approach reduces the machine setup time, and all system settings are automatically detected and optimised. 
The contribution of this paper can be summarised as follows:
\begin{itemize}
    \item A system setup for self optimisation and automatic code generation with an upstream and downstream unit.
    \item Illustration of the possibilities of action space adaptation in redundant priority, sequence arrangement and performance derivation. 
    \item Presentation of the interaction between EA, simulation, evaluation and code derivation in the intermediate unit.
    \item Implementation of the system setup in an industrial control process environment with a MOOP.
    \item Comparison of interactive EA adaptation with a combined sum weighting method by system knowledge. 
\end{itemize}
This paper reviews existing literature in~\ref{sec:Related_Work} and gives an overview  of EA in~\ref{sec:Application_of_Evolutionary_Algorithm_for_Process_Optimisation}. 
In~\ref{subsec:Process_specification_and_System_overview} the method of self optimisation and information processing is shown. 
Results and implementation are presented in~\ref{sec:Experiments_and_Results} and a summary is given in~\ref{sec:Conclusion}.

\section{Related Work}
\label{sec:Related_Work}
New standards have been set in the implementation of technical and scientific approaches. 
This digital transformation, also known as Industry $4.0$~\cite{Ghobakhloo}, includes decentralisation, automation, smart factory, augmented and virtual reality, to name a few. 
IoT technology enables wireless exchange~\cite{Li} for further applications~\cite{Chaudhary}\cite{Zarzo}.
Based on these principles, the introduction of Artificial Intelligence (AI) in industry has been enabled in applications such as fault diagnosis of rotating machinery~\cite{Liu} or path planning for underwater vehicles~\cite{Alvarez}. 
The aspect of logical understanding in relation to process flows is becoming increasingly important. In~\cite{Schwung} a game theory based learning approach is used to optimise an operational plant. This application area is considered and used as the basic scope for the EA investigation.
Although different approaches in Machine Learning (ML) provide solutions to different problems, EA is been particularly prominent in this area.
EA belongs to the stochastic, metaheuristic optimisation approaches~\cite{MAIER}, which are inspired by biological processes and physical conditions~\cite{QIAO2022}. 
Influences such as decision strategies~\cite{Kaim}, quantum computing~\cite{Deng} and many others have led to the development of diverse variations and characteristics of individual strategies and approaches in this field. 
Essential classifications can be made between Genetic Algorithm (GA) where optimisation is carried out based on tendency and combination, Genetic Programming (GP) where a specific structure is adapted and Evolution Strategies (ES) which are mainly based on numerical vector optimisation~\cite{Slowik}\cite{Deng2022}. 
The implementation of this approach is based on the principles of GA for inheritance and the strategies of ES.
The great popularity of EA can be seen in the application areas of MOOPs. 
MOOPs deals with the minimisation or maximisation of multiple transfer functions that are pursued as objectives.
This approach takes these advantages and applies them to industrial problems. 
Due to the fact of mutual correlation, different results can be obtained by searching for solutions. 
Convergent solutions are described as a Pareto optimum, which together are considered as a Pareto front~\cite{Kalyanmoy}. 
Given the complexity of multiple objective functions and Pareto optima, there are different ways to proceed depending on the optimisation and decision process.
In~\cite{Veldhuizen} a distinction is made between three Preference Articulation (PA), which lead to different solution approaches.
A distinction is made between prior, progressive and posterior PA.
The application of the prior PA is shown in~\cite{Schwung2}, where a system with multiple goal dependencies is continuously optimised by a single weighted reward function.
Extensive knowledge of the system behaviour is used to constrain the solution space.
A progressive PA, also known as interactive optimisation, operates after a continuous change of solution finding and decision making.
Depending on the focus, different approaches and strategies are known for the implementation~\cite{SHIN}. 
The approach presented, based on the strongest survivors, achieves stable learning outcomes. 
A posteriori PA such as~\cite{INGHELS} is discarded for this study, as the system optimisation would only be considered in one sub-area.

\section{Application of Evolutionary Algorithm for Process Optimisation}
\label{sec:Application_of_Evolutionary_Algorithm_for_Process_Optimisation}
\begin{figure}[htbp]
\centerline{\includegraphics[scale=0.5]{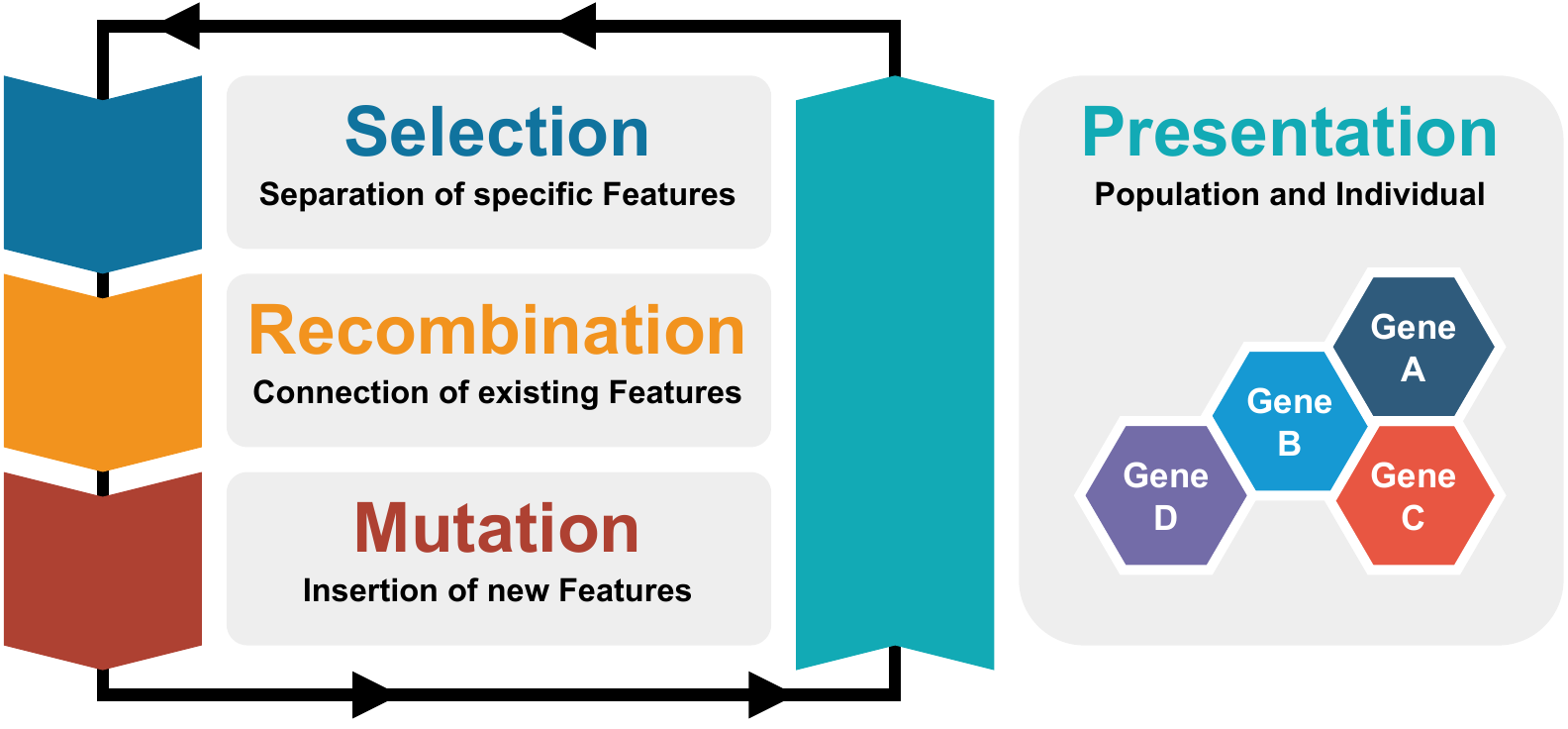}}
\caption{Structure of EA. A cyclical process of evolution by selection, recombination and mutation. The genetics of individuals adapt through this process and change the order in the population.}
\label{fig:Evolution_Process}
\end{figure}

The use of EA which combines the advantages of stochastic and metaheuristic optimisation is proposed. 
Continuous adaptation and optimisation is achieved through principles inspired by the processes of nature and evolution Figure~\ref{fig:Evolution_Process}. 

\subsection{Population and Individual Structure}
\label{subsec:Population_and_Individual_Structure}
Exchange and modification are fundamental stages in the structure of EA. 
The knowledge and understanding created must be stored centrally.
For this purpose, a set $P$ are created as a population with the cardinality $|P|=p$. 
This represents the basic unit containing $p$ individuals as sets $I$, where $I \in \{ x~|~P(x)\}$. 
Each of these individuals contains $E(m)$ specific elements and $T(n)$ tuples of information called genes.
Throughout the process, the structure of the population and the individuals is identical.

\subsection{Selection Process}
\label{subsec:Selection_Process}
An important aspect of the evolutionary process is the selection of the individual $I$.
The individual selection is done on the specific elements and tuple information and defines the restriction order $R$. 
As a result, a pre-order $(P, R)$ is generated from the population over the individual elements $E(m)$. 
Based on a minimisation $\forall~y \in P:I \le y$ or a maximisation $\forall~y \in P:I \ge y$, a specific selection of individuals $I \in P$ with the desired characteristics is made. 
The excluded individuals $y$ from the population $P$ can be selected at a later stage, provided that the condition is met.

\subsection{Recombination Method}
\label{subsec:Recombination_Method}
Existing information can be assumed by recombination. 
This method chooses the favourite elements $a$ from the individual specific tuples $T(n)$ described by $a \in \{x_{1}, x_{2}, \dots, x_{n}\}$. 
All selected elements and information contained in $a$ are combined into the new tuple $M$ with $\{x|\exists~a \in M: x \in a\}$. 
The chosen element information is discovered by a new single structure $I_{M}$. 
By aspiring to the initial structure, there is an integration of the recombining individual $I_{M}$ into the population $P$, where $I_{M} \in P$.

\subsection{Mutation Technique}
\label{subsec:Mutation_Technique}
Explicit modification can be achieved by the technique of mutation. 
The specific elements $a$ to be modified are selected from the individual specific tuples $T(n)$ with $a \in \{x_{1}, x_{2}, \dots, x_{n}\}$. 
Chosen elements and information $a$ are mutated in the range of defined element boundaries of $E(m)$ so that $a \in \{x_{1~new}, x_{2~new}, \dots, x_{n~new}\}$ results. 
The retained elements and the mutated set are combined in $M$ with $\{x~|~\exists~a \in M: x \in a\}$. 
By retaining the described structure, $I_{M} \in P$, where $I_{M}$ is a mutant individual that can be integrated into the population $P$.

\section{Process specification and System overview}
\label{subsec:Process_specification_and_System_overview}
\begin{figure}[htbp]
\centerline{\includegraphics[scale=0.34]{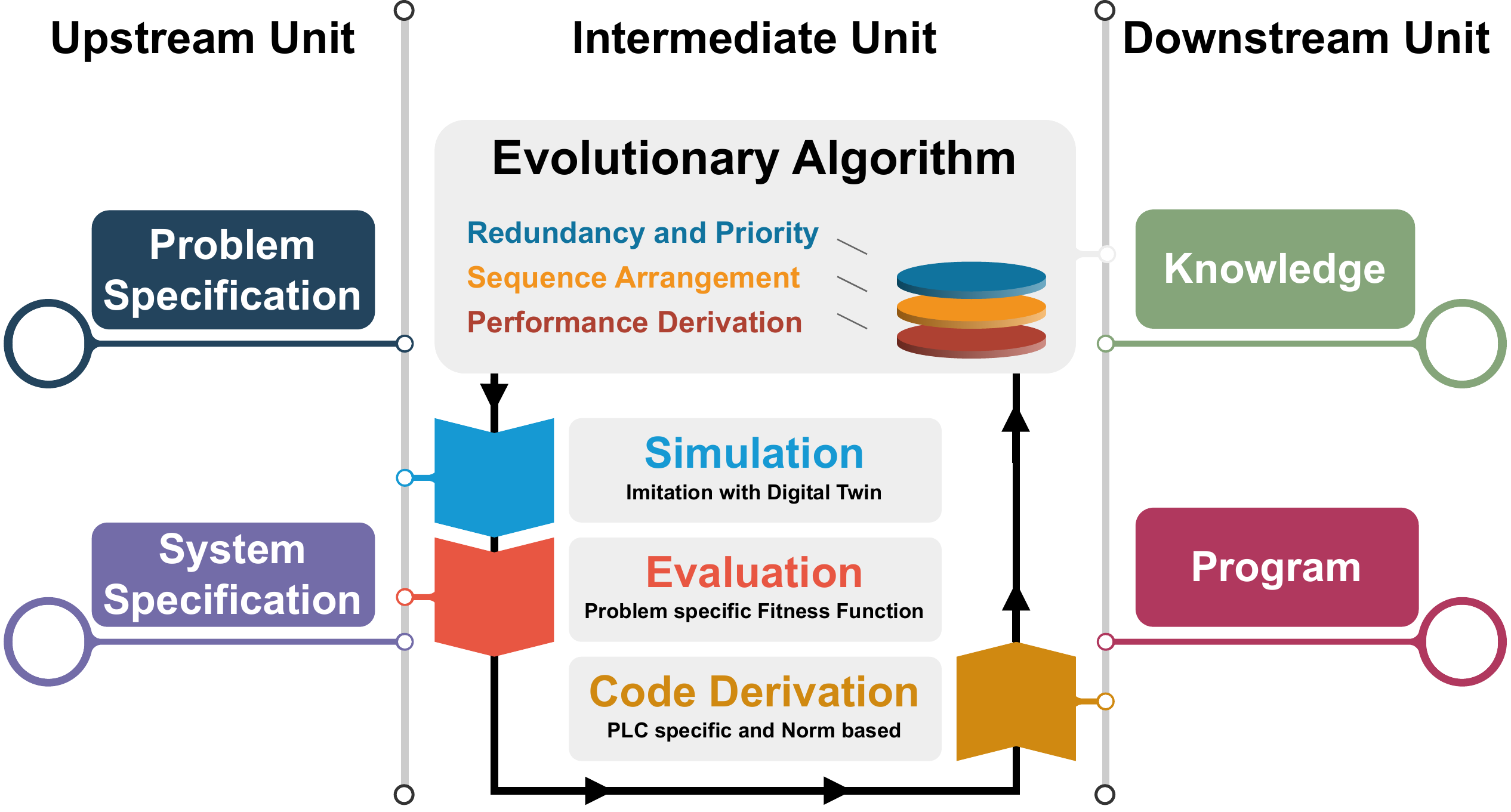}}
\caption{Illustration of the proposed framework for self optimisation and automatic code generation. The serial interface flexibly connects the system to the customised problem representation and to the output representation.}
\label{fig:Concept}
\end{figure}

To solve the problem of self optimisation and automatic code generation the proposed  system setup can be categorised into an upstream, an intermediate and a downstream unit. An overview of the developed system setup is presented in Figure~\ref{fig:Concept} and shows the general arrangement.

The developed setup has connectable interfaces on the primary side to the specific problem characteristics and the specifications of the system environment and is used as an upstream unit. 
The intermediate unit is centrally located in the system setup and consists of an optimisation algorithm, a replication of the system environment implemented as a DT, an evaluation unit with a problem specific fitness function and a code derivation unit.
This is contrasted by the downstream unit with the output of the system knowledge analyser and a programme decoder unit head on the secondary side.
This arrangement makes the proposed approach flexible and allows it to be sequentially extended or modified in the case of alternative starting points. 
The consideration of the interaction of the system setup is passive, as the evaluation cycle interacts with the DT of the system environment.
Derivation of the solution into a standard PLC program according to IEC~61131, IEC~61499 or IEC~61512 specifications is possible. 
After the optimisation process, the knowledge gained and the generated code can be actively used for the specific system. 
The scenario created in this way runs passively through a continuous cyclical optimisation cycle between simulation and observation based on an evaluation followed by an active solution output.

\subsection{Information Processing by Individuals and Populations}
\label{subsec:Information_Processing_by_Individuals_and_Populations}
\begin{figure}[htbp]
\centerline{\includegraphics[scale=0.7]{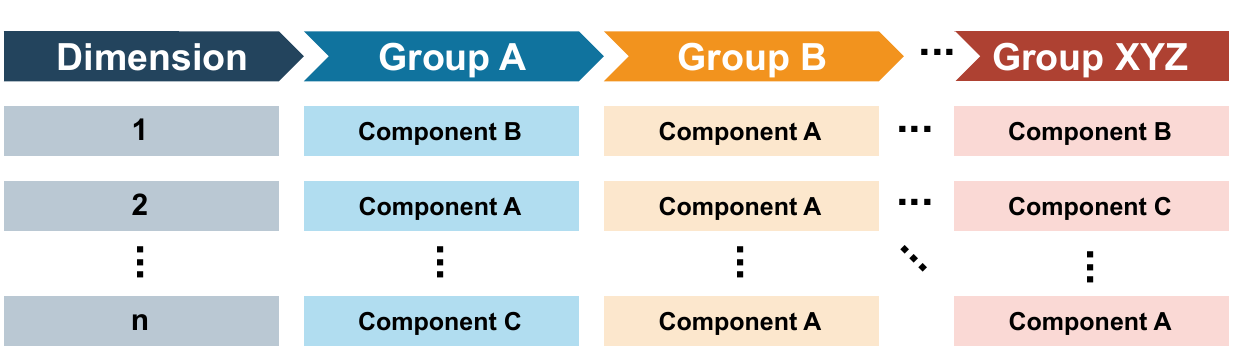}}
\caption{Illustration of the sequential individual structure. The arrangement of the group components and the order of the dimensions are varied during the evolutionary process.}
\label{fig:Sequential_Individual_Structure}
\end{figure}

A central aspect that is essential to the application presented is the arrangement and presentation of information. 
This is achieved through the use of individuals and populations, which play a central role in the EA. 
Embedded in this representation are all the necessary system parameters, objective and constraint functions and constraints required to influence and control the process.
For PLCs, standardised methods can be implemented based on function blocks, instruction lists, ladder diagrams, structured control language or sequential control.
The basic principles of IEC~61131 and IEC~61499 are followed~\cite{DIN_EN_61131}\cite{DIN_EN_61499}.
This provides a flexible, standardised structure for defining PLC program code.
The measures taken allow for a cross-hierarchical approach in which it is possible to implement different code generation strategies, defined as individuals in the case of EA.
The code structure is considered in sequences where the composition determines the behaviour. 
The focus is on redundancy and the definition of priorities.
The action space can be defined to include or exclude a relevant part depending on the action. 
This results in an optimisation of relevant sequences that provide the context of the control logic.
Another consideration is the arrangement of the sequences. 
Differences are recorded in the order of their activation and checked against simplified redundant parts to be replaced. 
By identifying the hierarchical cycle and resolving complex structures, a higher level sequence structure is generated. 
By combining the two variants, another representation of the general performance requirement can be derived, which can be used as a procedure instruction.
A possible extension is to derive a general procedural instruction that expresses the behaviour and knowledge of the system behaviour.
This provides a better understanding of the interaction between the problem and the solution in the context of the bigger picture. 
Possible representations that affect the implementation of individuals and populations are system specific and can be adapted specifically to the approach.
For the implementation, a table structure is defined for the individual representation Figure~\ref{fig:Sequential_Individual_Structure}.

The arrangement creates a flexible representation of integrated components, operating modes and a clear connection between dependencies.
Changes can be made to component groups to incorporate different effects as well as the number of dimensions displayed to vary in complexity.

\subsection{Self Optimisation through the Intermediate Unit}
\label{subsec:Self_Optimisation_through_the_Intermediate_Unit}
Special attention must be paid to the implementation of the intermediate unit. 
The optimisation process is implemented by an integrated EA that continuously explores the search space by varying and independently adapting possible solutions.
In order to prevent the system from going to a local extreme during the optimisation, the functions of the EA are defined according to the following focus. 
The selection function is designed in such a way as to select from the population those individuals that are best suited to the parameter of interest. 
This selection is used for the mutation function. 
By comparison, the genetic characteristics of the best performing individuals are selected and recombined to create a new individual.  
To allow for new features in the evolutionary process, the mutation function randomly selects genetic characteristics from the individual and mutates them. 
The clear structure of the algorithm ensures a simple and efficient learning strategy.
The optimisation by the EA is defined by the determination of the action space and the fitness function. 
Possibilities of action space representation have been shown and can be extended to other applications. 
Depending on the problem definition, a fitness function has to be defined to evaluate the active interaction with the DT of the adapted system. 
This represents the goal of the adaptation, which can be further complemented by specific parameters.
To generate usable solutions from the weighted evolution, a code derivation for generation is integrated. 
The described implementations are called cyclically in a loop. 
At the end of the cycle, the developed individuals are ranked in the population with the weighted assessment results.
The design of the optimisation runs can be defined based on the quality criteria of the fitness function
The proposed intermediate unit enables automatic code generation through self optimisation and generates system knowledge and program code.

\section{Experiments and Results}
\label{sec:Experiments_and_Results}
\begin{figure}[htbp]
\centerline{\includegraphics[scale=0.6]{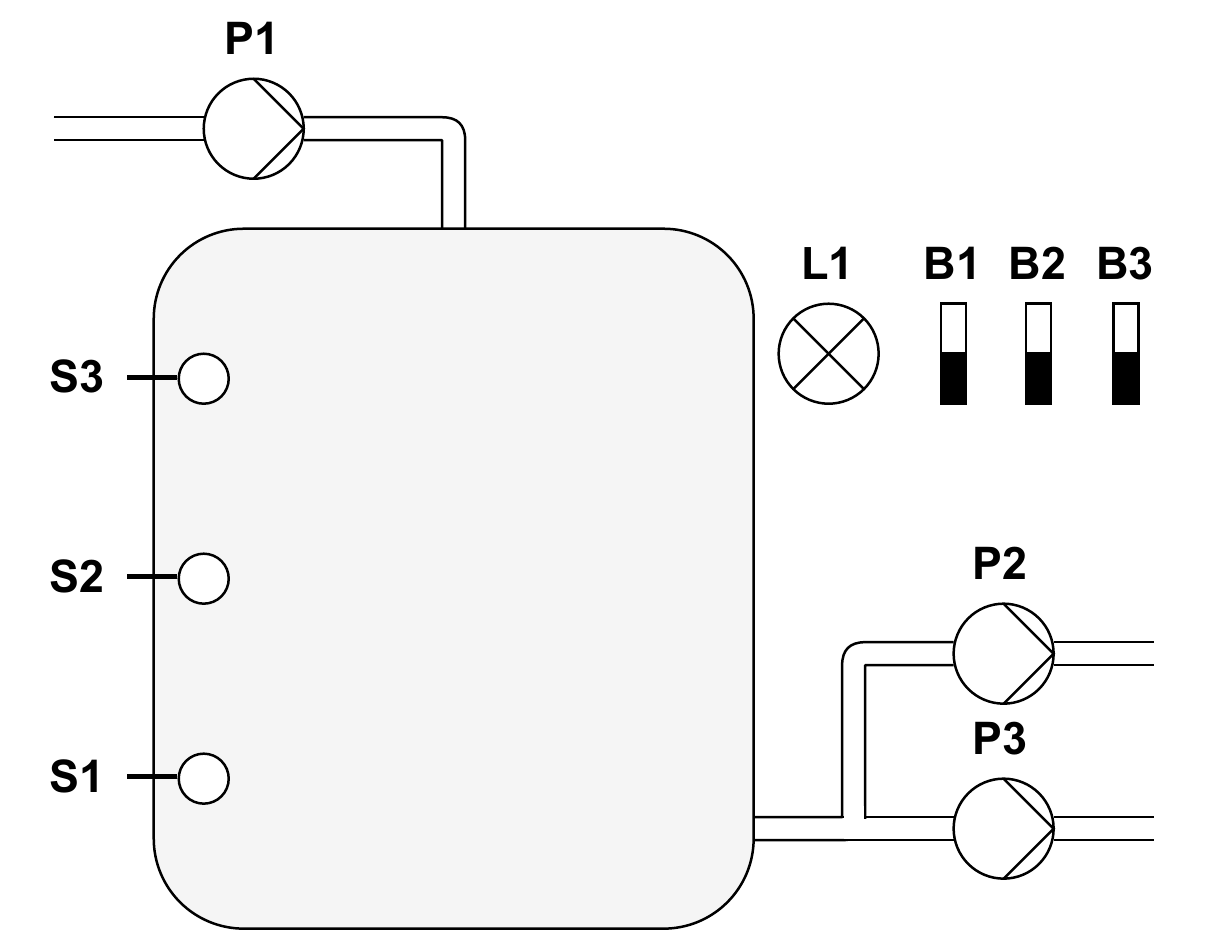}}
\caption{Schematic layout of the liquid laboratory station with identification of individual actuators and sensors. 
The level in the tank $T$ is raised by pump $P_{1}$ and lowered by pump $P_{2}$ and pump $P_{3}$.}
\label{fig:Laboratory_Plant}
\end{figure}

The analysis takes place in a liquid laboratory plant as shown in Figure~\ref{fig:Laboratory_Plant}. 
Based on the customised system characteristics, the liquid laboratory plant is considered as a multi-objective optimisation problem, which can be described by the general form according to Eq.~\ref{eq:multi-objective_optimisation_problem}.
\begin{equation}
    \begin{aligned}
        \text{min or max} && f_{m}(\mathbf{x}), && m &= 1, 2, \dots, M; \\
        \text{subject to} && g_{j} > 0, && j &= 1, 2, \dots, J; \\
        \text{} && h_{k} = 0, && k &= 1, 2, \dots, K; \\
        \text{with} && x_{i}^{L} \le x_{i} \le x_{i}^{U}, && i &= 1, 2, \dots, n; \\
    \end{aligned}
    \label{eq:multi-objective_optimisation_problem}
\end{equation}
The consideration behaviour is related by $\mathbf{x}$, a vector with $n$ variables $\mathbf{x} = (x_{1}, x_{2}, \dots, x_{n})^{T}$ and the constraint functions $g_{j}$ and $h_{k}$, associated by $J$ inequality and $K$ equality constraints. 
A variable restriction takes place in the region of the lower boundary $x_{i}^{L}$ and the upper boundary $x_{i}^{U}$. 
With these restrictions, the independent objective functions $f$ form a variable space $D$. 
Depending on $\mathbf{x} \in D$, the variable vector is called a feasible solution or an infeasible solution by $\mathbf{x} \in D$. 
The result of $M$ multiple objective functions $f_{m}$ is a multidimensional space called the objective space $Z$, where for every feasible solution $f_{x}$ there exists a point $z$ where $f_{\mathbf{x}} = \mathbf{z} = (z_{1}, z_{2}, \dots, z_{n})^{T}$ \cite{Kaisa}\cite{Kalyanmoy}. Due to internal controversies between the objective functions, a Pareto dominance occurs between different vectors which is resulting in a dominated ratio of $\mathbf{x}^{\prime}~\preceq~\mathbf{x}$. 
For $\mathbf{x}~\in~D$, the objective functions lead to an efficient solution where $\{F(x) = (f_{1}(x), f_{2}(x), \dots, f_{m}(x))\}$ dominates over $F(x^{\prime})$. 
Starting from an efficient solution with the set $\mathbf{x}$, a Pareto efficient solution can be defined as the set $\mathcal{P}$ by $\{x \in D|\lnot~\exists~x^{\prime} \in D:F(x^{\prime})\preceq F(x)\}$. 
The interdependence of the objective functions creates several partial possibilities $\mathcal{PF}$, known as the Pareto front, described by $\{ F(x)=(f_{1}(x), f_{2}(x), \dots, f_{m}(x)~|~x \in \mathcal{P}\}$ \cite{Veldhuizen}.

\subsection{Liquid Laboratory Plant}
\label{subsec:Liquid_Laboratory_Plant}
The system is characterised by a central tank $T$ which is filled and emptied by pipes. 
The specific volume flow depends on the respective activation of the pumps $P_{1}$, $P_{2}$ and $P_{3}$.  
An actual filling level of the tank system is detected by sensors $S_{1}$, $S_{2}$ and $S_{3}$. 
The operation of the laboratory plant is differentiated by different functional states. 
The laboratory plant can be operated in automatic or manual mode which is selected by the switch $B_{1}$. 
In automatic mode, the pump activation for filling and emptying is based on the current level. 
Automatic operation is indicated by a control light $L_{1}$. 
In manual mode, the fill and drain pumps can be controlled separately by a plant manager using the $B_2$ and $B_3$ keys. 
The liquid laboratory process is centrally controlled by a PLC.
The programme code is structured according to the EN~61131~\cite{DIN_EN_61131} standard and uses the listed sensors and actuators.

In the table~\ref{tab:Components_of_the_Laboratory_Plant}, a classification of the components of the laboratory plant is made between inputs and outputs. 
System states and output components are adjusted based on the input components.
Process optimisation is performed according to maximum transport $P_{trans}$, minimum energy $P_{energy}$ and code structure $P_{code}$. 
The function and operating structure of the liquid laboratory station is divided into automatic and manual modes. 

\begin{table}[htbp]
  \centering
  \caption{List of components of the liquid laboratory plant. Classification into passive input and active output parts.}
    \begin{tabular}{|c|c|c|c|}
    \hline
    \textbf{Label} & \textbf{Type} & \textbf{Unit} & \textbf{Description} \\
    \hline
    $S_{1}$    & Input & Sensor & Detect Level Position 1 \\
    $S_{2}$    & Input & Sensor & Detect Level Position 2 \\
    $S_{3}$    & Input & Sensor & Detect Level Position 3 \\
    $B_{1}$    & Input & Switch & Automatic Manuel Mode \\
    $B_{2}$    & Input & Button & Manual Filling \\
    $B_{3}$    & Input & Button & Manual Empty \\
    \hline
    $P_{1}$    & Output & Pump  & Active Pump 1 \\
    $P_{2}$    & Output & Pump  & Active Pump 2 \\
    $P_{3}$    & Output & Pump  & Active Pump 3 \\
    $L_{1}$    & Output & Light & Active Light 1 \\
    \hline
    \end{tabular}
  \label{tab:Components_of_the_Laboratory_Plant}
\end{table}
\begin{itemize}
    \item Automatic filling and emptying is performed by pumps $P_{1}$, $P_{2}$ and $P_{3}$. Activation depends on the level detected by sensors $S_{1}$, $S_{2}$ and $S_{3}$. 
    The system is in automatic mode when the $B_1$ gives a high signal. 
    The mode is indicated by the control light $L_{1}$.
    \item Manual filling is achieved by a low signal from the switch $B_1$ and a high signal from the fill button $B_{2}$. This activates the pump $P_{1}$.
    \item Manual emptying of the tank is performed by pumps $P_{2}$ and $P_{3}$. They are activated by a low signal from the switch $B_1$ and a high signal from the empty button $B_3$.
\end{itemize}
Delayed activation and deactivation is caused by the system hysteresis.

\subsection{System Transfer and Implementation}
\label{subsec:System_Transfer_and_Implementation}
The liquid laboratory plant described is considered in automatic mode. 
Optimisation is performed within the mathematical and technical limits of the system. 
The obtained dependence of the MOOP is solved after prior PA with a sum weight method and a progressive PA where an interactive optimisation takes place.
Conditions that do not satisfy the boundary condition are excluded. 
Based on the objectives of the liquid laboratory station process $P_{trans}$, $P_{energy}$ and $P_{code}$, the optimisation approach is implemented.
For the previous PA, the current objective values of transport $f_{1}$, energy consumption $f_{2}$ and programme code structure $f_{3}$ are combined into a separate function according to the sum weighting method, which is used as the fitness function of the EA with

\begin{equation}
\textstyle
    F_{fit}= \Bigl( \frac{1}{1+\alpha_{1} (P_{trans}-f_{1})}+\frac{1}{1 + \alpha_{2} f_{2}} + \frac{1}{1+\alpha_{3} (P_{code}-f_{3})} \Bigl).
    \label{eq:fitness_function}
\end{equation}

The resulting fitness function $F_{fit}$ can be optimised in a maximum that is influenced by the parameters $\alpha_{1}$, $\alpha_{2}$ and $\alpha_{3}$.
An understanding of the system behaviour is required to define the parameters for the fitness function. 
Created individuals that meet these requirements are stored in the EA population.
In contrast to this approach is the technical realisation of a progressive articulation of the EA. 
In this approach, all individuals with the relevant genes and outcomes are stored in the population.
Through selection, recombination and mutation of the best individuals, continuous genetic transfer creates an evolving optimisation throughout the process.

\subsection{Performance Presentation}
\label{subsec:Performance_Presentation}
\begin{figure}[htbp]
\centerline{\includegraphics[scale=0.7]{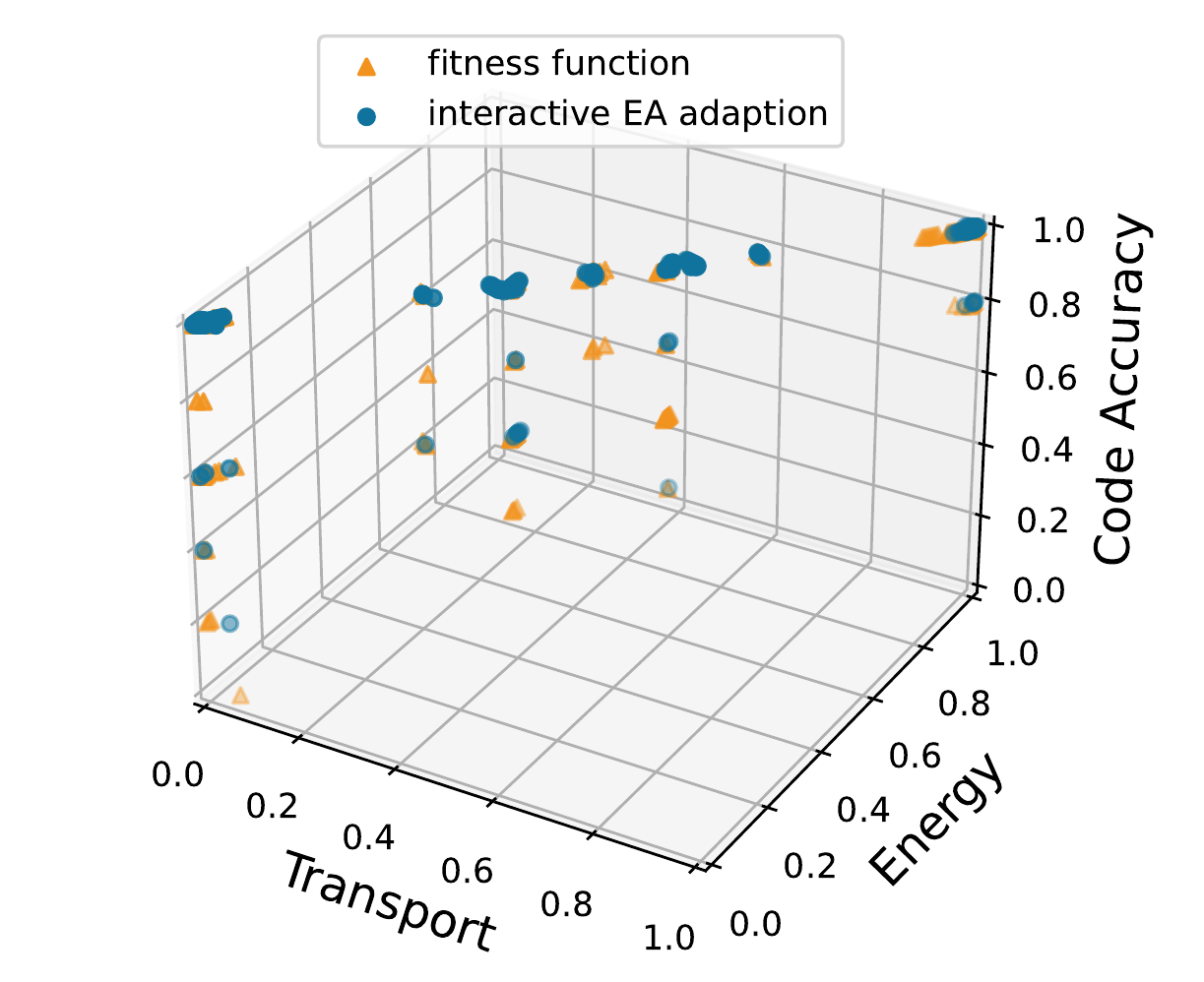}}
\caption{Comparison of the prior PA with the fitness function (orange triangles) and the progressive PA with genetic transfer (blue circles). The combination of transport and energy leads to parallel positioning in the target space.}
\label{fig:Objective_Space}
\end{figure}

The specific activation behaviour of the system components influences the process of the liquid laboratory station and changes the process flow. 
By transforming the MOOP into the objective space $D$, the influences and effects of the coupled objective functions can be represented. 
A comparison like Figure~\ref{fig:Objective_Space} shows the dependencies of the different objective functions.

From the perspective of system and process knowledge, the solutions in the search space are limited by the fitness function. 
By promoting and shaping specific solutions, a continuous learning approach is achieved throughout the optimisation process. 

\begin{figure}[htbp]
\centerline{\includegraphics[scale=0.55]{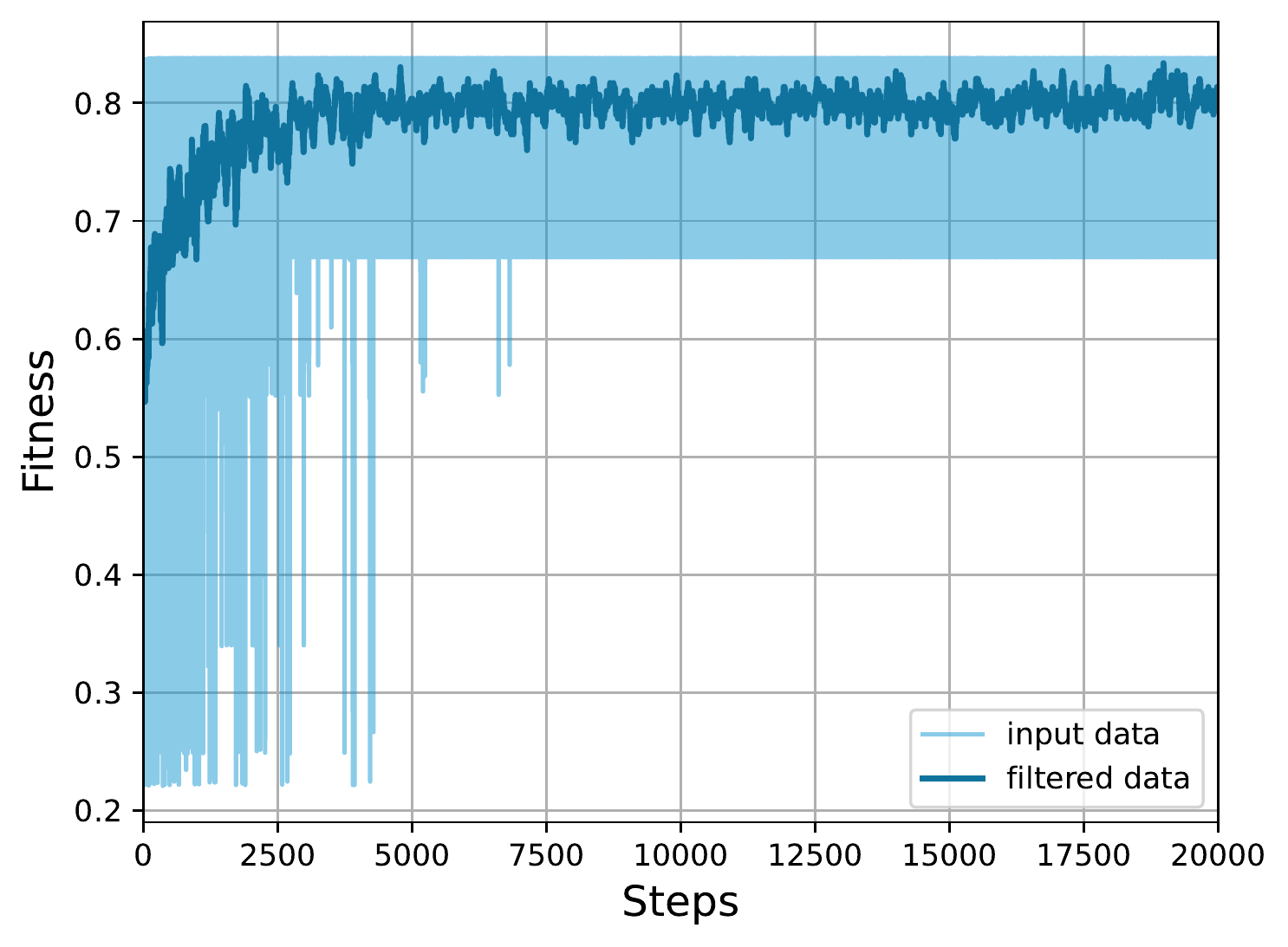}}
\caption{Demonstration of the increasing experience success of the prior PA with the fitness function. A stable state is achieved by weighting and balancing the different objective functions.}
\label{fig:Experience_success}
\end{figure}

\subsection{Decoding and Code Division}
\label{subsec:Decoding_and_Code_Division}
Based on the performance presentation of the MOOP, a possible individual solution can be selected from the prior or progressive PA. 
Table~\ref{tab:Code_Table_Structure} represents an individual~$I$ which is evaluated at the DT of the liquid laboratory plant.

\begin{table}[htbp]
  \centering
  \caption{The genetic structure of an individual represents the activation composition. The EA modifies content and information through selection, recombination and mutation.}
    \begin{tabular}{|c|c|c|c|c|c|}
    \hline
    \textbf{Line:} & \textbf{Output:} & \textbf{Input:} & \textbf{Operator:} & \textbf{Negation:} & \textbf{Mode:} \\
    \hline
    0     & P1    & S3    & =     & ¬     & A \\
    1     &       & B1    & $\land$     &       & A \\
    2     & P2    & S2    & =     &       & A \\
    3     &       & B1    & $\land$     &       & A \\
    4     & P3    & S1    & =     &       & A \\
    5     &       & B1    & $\land$     &       & A \\
    6     & L1    & B1    & =     &       & A \\
    \hline
    \end{tabular}%
  \label{tab:Code_Table_Structure}%
\end{table}%

The outputs, inputs, operators, negations and operating modes of the liquid laboratory station are implemented as component groups in the table representation.
The system logic for process control of the liquid laboratory station can be derived by calling up the structured arrangement of the individual~$I$ by the decoding unit.
A structured representation is realised, as is the case with programming languages such as structured control language or instruction lists.

\begin{equation}
    \begin{aligned}
        P{1} &= \bar{S{3}}~\text{and}~B_{1}; \\
        P{2} &= S{2}~\text{and}~B_{1}; \\
        P{3} &= S{1}~\text{and}~B_{1}; \\
        L{1} &= B_{1};
    \end{aligned}
    \label{eq:program_code_logic}
\end{equation}

The solution shown represents the sensor and actuator connection in a program code structure for the automatic operation of the liquid laboratory plant. 

\section{Conclusion}
\label{sec:Conclusion}
This paper proposes a novel approach to self optimisation with automatic code generation. 
It is based on EA and can be integrated into PLC based control processes.
The basics of selection, recombination and mutation in EA are explained and presented. 
The framework developed is divided into upstream, intermediate and downstream units and can be specifically parameterised. 
Continuous exploration takes place through interaction with a DT and an evaluation function.
System knowledge and solution proposals are optimised through redundancy and priority, sequencing and performance derivation in a customisable structural representation.
The approach is implemented and evaluated in the liquid laboratory plant to demonstrate the complexity of MOOP in real world applications. 
Results and a programme code derivation are presented.

Future work will investigate knowledge transfer and search space specification by multi-agents. 
One-to-one traceability should be integrated into the evolution process using bijective functionalities. 
By representing the population with neural networks, the potential and possible applications are to be expanded.

\bibliographystyle{IEEEtran}
\bibliography{reference.bib}

\begin{thebibliography}{10}
\providecommand{\url}[1]{#1}
\csname url@samestyle\endcsname
\providecommand{\newblock}{\relax}
\providecommand{\bibinfo}[2]{#2}
\providecommand{\BIBentrySTDinterwordspacing}{\spaceskip=0pt\relax}
\providecommand{\BIBentryALTinterwordstretchfactor}{4}
\providecommand{\BIBentryALTinterwordspacing}{\spaceskip=\fontdimen2\font plus
\BIBentryALTinterwordstretchfactor\fontdimen3\font minus
  \fontdimen4\font\relax}
\providecommand{\BIBforeignlanguage}[2]{{%
\expandafter\ifx\csname l@#1\endcsname\relax
\typeout{** WARNING: IEEEtran.bst: No hyphenation pattern has been}%
\typeout{** loaded for the language `#1'. Using the pattern for}%
\typeout{** the default language instead.}%
\else
\language=\csname l@#1\endcsname
\fi
#2}}
\providecommand{\BIBdecl}{\relax}
\BIBdecl

\bibitem{Kagermann}
H.~Kagermann and W.~Wahlster, ``\BIBforeignlanguage{eng}{Ten years of industrie
  4.0},'' \emph{\BIBforeignlanguage{eng}{Sci}}, vol.~4, no.~26, pp. 26--, 2022.

\bibitem{Khan}
M.~U.~K. Khan and C.-M. Kyung, ``\BIBforeignlanguage{eng}{Poisson mixture model
  for high speed and low-power background subtraction},'' in
  \emph{\BIBforeignlanguage{eng}{Smart Sensors and Systems}}.\hskip 1em plus
  0.5em minus 0.4em\relax Cham: Springer International Publishing, 2020, pp.
  1--23.

\bibitem{Chaudhary}
V.~Chaudhary, A.~Kaushik, H.~Furukawa, and A.~Khosla, ``Review—towards 5th
  generation ai and iot driven sustainable intelligent sensors based on 2d
  mxenes and borophene,'' \emph{ECS Sensors Plus}, vol.~1, no.~1, p. 013601,
  apr 2022.

\bibitem{Zarzo}
M.~Zarzo, A.~Perles, R.~Mercado, and F.-J. García-Diego,
  ``\BIBforeignlanguage{eng}{Multivariate characterization of temperature
  fluctuations in a historical building using energy-efficient iot wireless
  sensors},'' \emph{\BIBforeignlanguage{eng}{Sensors (Basel, Switzerland)}},
  vol.~21, no. 7795, pp. 7795--, 2021.

\bibitem{Singh}
M.~Singh, E.~Fuenmayor, E.~Hinchy, Y.~Qiao, N.~Murray, and D.~Devine, ``Digital
  twin: Origin to future,'' \emph{Appl. Syst. Innov}, vol.~4, p.~36, 05 2021.

\bibitem{Kalyanmoy}
K.~Deb, \emph{\BIBforeignlanguage{eng}{Multi-objective optimization using
  evolutionary algorithms}}, ser. Wiley Interscience ser. syst. optim.\hskip
  1em plus 0.5em minus 0.4em\relax Chichester [u.a: Wiley, 2002.

\bibitem{Ghobakhloo}
M.~Ghobakhloo, ``Industry 4.0, digitization, and opportunities for
  sustainability,'' \emph{Journal of Cleaner Production}, vol. 252, p. 119869,
  2020.

\bibitem{Li}
S.~Li, L.~D. Xu, and S.~Zhao, ``The internet of things: a survey,''
  \emph{Information Systems Frontiers}, vol.~17, no.~2, pp. 243--259, Apr 2015.

\bibitem{Liu}
R.~Liu, B.~Yang, E.~Zio, and X.~Chen, ``Artificial intelligence for fault
  diagnosis of rotating machinery: A review,'' \emph{Mechanical Systems and
  Signal Processing}, vol. 108, pp. 33--47, 2018.

\bibitem{Alvarez}
A.~Alvarez, A.~Caiti, and R.~Onken, ``Evolutionary path planning for autonomous
  underwater vehicles in a variable ocean,'' \emph{IEEE J. Ocean. Eng.},
  vol.~29, no.~2, pp. 418--429, 2004.

\bibitem{Schwung}
D.~Schwung, J.~Reimann, A.~Schwung, and S.~Ding, \emph{Smart Manufacturing
  Systems: A Game Theory based Approach}.\hskip 1em plus 0.5em minus
  0.4em\relax Springer Nature Switzerland AG 2020, 01 2020.

\bibitem{MAIER}
H.~Maier, S.~Razavi, Z.~Kapelan, L.~Matott, J.~Kasprzyk, and B.~Tolson,
  ``Introductory overview: Optimization using evolutionary algorithms and other
  metaheuristics,'' \emph{Environmental Modelling \& Software}, vol. 114, pp.
  195--213, 2019.

\bibitem{QIAO2022}
K.~Qiao, J.~Liang, K.~Yu, M.~Yuan, B.~Qu, and C.~Yue, ``Self-adaptive resources
  allocation-based differential evolution for constrained evolutionary
  optimization,'' \emph{Knowledge-Based Systems}, vol. 235, p. 107653, 2022.

\bibitem{Kaim}
A.~Kaim, B.~Bartkowski, N.~Lienhoop, C.~Schröter-Schlaack, M.~Volk, and
  M.~Strauch, ``Combining biophysical optimization with economic preference
  analysis for agricultural land-use allocation,'' \emph{Ecology and Society},
  vol.~26, 03 2021.

\bibitem{Deng}
W.~Deng, H.~Liu, J.~Xu, H.~Zhao, and Y.~Song, ``An improved quantum-inspired
  differential evolution algorithm for deep belief network,'' \emph{IEEE Trans
  Instrum Meas}, vol.~69, no.~10, pp. 7319--7327, 2020.

\bibitem{Slowik}
A.~Slowik and H.~Kwasnicka, ``Evolutionary algorithms and their applications to
  engineering problems,'' \emph{Neural Computing and Applications}, vol.~32,
  no.~16, pp. 12\,363--12\,379, Aug 2020.

\bibitem{Deng2022}
W.~Deng, X.~Zhang, Y.~Zhou, Y.~Liu, X.~Zhou, H.~Chen, and H.~Zhao, ``An
  enhanced fast non-dominated solution sorting genetic algorithm for
  multi-objective problems,'' \emph{Information Sciences}, vol. 585, pp.
  441--453, 2022.

\bibitem{Veldhuizen}
D.~Veldhuizen and G.~Lamont, ``Multiobjective evolutionary algorithms:
  Analyzing the state-of-the-art,'' \emph{Evolutionary computation}, vol.~8,
  pp. 125--47, 02 2000.

\bibitem{Schwung2}
D.~Schwung, S.~Yuwono, A.~Schwung, and S.~Ding, ``Decentralized learning of
  energy optimal production policies using plc-informed reinforcement
  learning,'' \emph{Computers \& Chemical Engineering}, vol. 152, p. 107382, 05
  2021.

\bibitem{SHIN}
W.~S. Shin and A.~Ravindran, ``Interactive multiple objective optimization:
  Survey i—continuous case,'' \emph{Computers \& Operations Research},
  vol.~18, no.~1, pp. 97--114, 1991.

\bibitem{INGHELS}
D.~Inghels, W.~Dullaert, E.-H. Aghezzaf, and R.~Heijungs, ``Towards optimal
  trade-offs between material and energy recovery for green waste,''
  \emph{Waste Management}, vol.~93, pp. 100--111, 2019.

\bibitem{DIN_EN_61131}
``Din en 61131-3 speicherprogrammierbare steuerungen - teil 3:
  Programmiersprachen,'' Jun. 2014.

\bibitem{DIN_EN_61499}
``Funktionsbausteine für industrielle leitsysteme - teil 1: Architektur,''
  Sep. 2014.

\bibitem{Kaisa}
K.~Miettinen, \emph{\BIBforeignlanguage{eng}{Nonlinear multiobjective
  optimization}}, ser. International series in operations research \&
  management science 12.\hskip 1em plus 0.5em minus 0.4em\relax Boston [u.a:
  Kluwer Academic Publishers, 1999.

\end{thebibliography}

\end{document}